\documentclass[letterpaper]{article} 
\usepackage{aaai2026}  
\usepackage{times}  
\usepackage{helvet}  
\usepackage{courier}  
\usepackage[hyphens]{url}  
\usepackage{graphicx} 
\usepackage{natbib}  
\usepackage{caption} 
\usepackage{algorithm}
\usepackage{algorithmic}

\usepackage{newfloat}
\usepackage{listings}
\DeclareCaptionStyle{ruled}{labelfont=normalfont,labelsep=colon,strut=off} 
\floatstyle{ruled}
\newfloat{listing}{tb}{lst}{}
\floatname{listing}{Listing}
\usepackage{booktabs}
\usepackage{amsmath}
\usepackage{multirow}
\usepackage{multicol}
\usepackage{makecell}
\usepackage{subcaption}
\usepackage{fancyhdr}
\title{Egocentric Instruction-oriented Affordance Prediction via \\ Large Multimodal Model}
\author {
    Bokai Ji\textsuperscript{\rm 1 \dag *},
    Jie Gu\textsuperscript{\rm 2 \dag},
    Xiaokang Ma\textsuperscript{\rm 2},
    Chu Tang\textsuperscript{\rm 2},
    Jingmin Chen\textsuperscript{\rm 2},
    Guangxia Li\textsuperscript{\rm 1 \ddag}
}
\affiliations {
    \textsuperscript{\rm 1}Xidian University \quad
    \textsuperscript{\rm 2}Rightly Robotics\\
    jibokai@stu.xidian.edu.cn, gxli@xidian.edu.cn, \\
    \{jgu,xma,chu.tang,jingmin.chen\}@rightly.ai
}

\begin{document}

\maketitle
\thispagestyle{fancy}

\insert\footins{\noindent\footnotesize \dag \  Both authors contributed equally to this research.}
\insert\footins{\noindent\footnotesize * Interns at Rightly Robotics.}
\insert\footins{\noindent\footnotesize \ddag \ Corresponding Author.}

\begin{abstract}
Affordance is crucial for intelligent robots in the context of object manipulation. In this paper, we argue that affordance should be task-/instruction-dependent, which is overlooked by many previous works. That is, different instructions can lead to different manipulation regions and directions even for the same object. According to this observation, we present a new dataset comprising fifteen thousand object-instruction-affordance triplets. All scenes in the dataset are from an egocentric viewpoint, designed to approximate the perspective of a human-like robot. Furthermore, we investigate how to enable large multimodal models (LMMs) to serve as affordance predictors by implementing a ``search against verifiers'' pipeline. An LMM is asked to progressively predict affordances, with the output at each step being verified by itself during the iterative process, imitating a reasoning process. Experiments show that our method not only unlocks new instruction-oriented affordance prediction capabilities, but also achieves outstanding performance broadly. 
\end{abstract}
\section{Introduction}

Humans interact with a considerable number of objects every day. The accumulation of the interactions has endowed us with extensive knowledge on how most objects should be manipulated in various scenarios. This type of actionable knowledge is crucial for general-purpose embodied intelligent systems. Robot manipulation~\cite{DBLP:conf/icra/GengAGCYD23, DBLP:conf/humanoids/KokicSHK17, DBLP:conf/nips/NagarajanG20} relies on accurate perceiving and reasoning about where to act and in which direction to move for the successful execution of tasks. Predicting such visual affordances is a vital topic in both the robotics and the computer vision communities, with a wide range of applications.

Despite the significant work and progress within the community~\cite{DBLP:conf/eccv/JuHZZJX24, DBLP:conf/cvpr/BahlMCJP23, DBLP:conf/cvpr/LiuTM022, DBLP:conf/cvpr/0008SSJ24, DBLP:conf/cvpr/0008JSS23}, the basic question of how the affordance should be defined and represented remains controversial from our perspective. Instead of treating affordance as static and inherent properties of objects or object components, we argue that affordance should be dynamic and closely related to the purpose of interaction. That is, it should be instruction- or task- oriented. For example, the affordance of a drawer as push-able or pull-able depends on whether the task is to close or open it. Moreover, we notice that this field is advancing towards more intelligent and natural interactions with the environments. A considerable amount of research focuses on open-vocabulary manipulation~\cite{DBLP:conf/corl/HuangWZL0023}, egocentric visual understanding~\cite{DBLP:conf/cvpr/PeironePAA24, DBLP:journals/corr/abs-2410-11623}, and learning from videos of human interaction~\cite{DBLP:conf/cvpr/LiuTM022, DBLP:conf/cvpr/BahlMCJP23}. This helps bridge the gaps between visual affordance learning and robotic deployment.

We then introduce a new egocentric instruction-oriented affordance grounding task, which requires predicting contact regions and motion directions according to a given manipulation instruction/task. Alongside, a dataset is built for comprehensive studies and model capability benchmarks, named \underline{E}gocentric \underline{I}nstruction-oriented \underline{V}isual \underline{A}ffordance (EIVA). EIVA contains 14,861 egocentric scenes, with each scene features an object-instruction-affordance triplet, involving a total of 146 object-instruction combinations. The data sources include Ego4D~\cite{DBLP:conf/cvpr/GraumanWBCFGH0L22}, Epic-Kitchens~\cite{DBLP:journals/pami/DamenDFFFKMMPPW21} and HOI4D~\cite{DBLP:conf/cvpr/LiuLJLWSLFWY22}. The affordance annotations are first generated by an automated pipeline and then reviewed manually.

The core of the automated annotation pipeline is to lock onto the contacts between the hand and the object. Specifically, given an egocentric video of human interaction, an off-the-shelf hand-object detector~\cite{DBLP:conf/cvpr/ShanGSF20} first identifies the hand and the object, after which SAM2~\cite{DBLP:journals/corr/abs-2408-00714} is applied to extract the corresponding masks. For motion direction extraction, we prompt SpatialTracker~\cite{DBLP:conf/cvpr/Xiao0Z0PSZ24} with these masks to perform 3D point-level tracking. Principal Component Analysis (PCA) is applied to these tracked points within masks, and the motion direction can be approximated by the first principal component~\cite{witte2010applying}. In terms of contact regions, a key consideration is to avoid occlusion of the object from the hand. We sample points at the periphery of the hand mask as contact points, similar to~\cite{DBLP:conf/cvpr/BahlMCJP23}. These contact points are projected back onto earlier frames using homography until no interactions are detected (by~\cite{DBLP:conf/cvpr/ShanGSF20}). Subsequently, Gaussian Blur is applied to the projected contact points to obtain the affordance map. Finally, instructions are derived from timestamped narrations that describe actions in the form of $\langle verb \rangle \ the \ \langle noun \rangle$.

The introduced open-vocabulary instruction-oriented affordance prediction task in real-world scenarios is quite challenging. The model requires three key abilities: 1) fine-grained grounding various objects in complex scenes, regardless of their varying appearances, shapes, and sizes; 2) powerful multimodal understanding capabilities; 3) reasoning about \textit{How} and \textit{Where} to manipulate to complete tasks based on rich world knowledge. Training models with these capabilities and good generalization is very difficult due to the scarcity of large-scale data.

The large multimodal model (LMM) appears to fulfill the aforementioned capability requirements and holds great potential to complete the proposed task. Recent advancements have demonstrated that LMMs are endowed with extensive world knowledge~\cite{DBLP:conf/corl/HuangWZL0023}, including how to manipulate objects to accomplish tasks. These models also exhibit remarkable generalizability, suggesting that unlike previous approaches~\cite{DBLP:conf/cvpr/ChuangL0F18}, they are unlikely to struggle with handling unseen objects and novel affordances. We evaluate the closed-source model GPT-4o~\cite{gpt4o} and the open-source model Qwen2.5-VL~\cite{qwen25-vl} on the proposed dataset. As expected, the results show that LMM indeed has a certain level of affordance prediction ability.

During this process, we observe that the visual grounding capability of LMMs is still not satisfactory, probably due to the complex scene and egocentric viewpoint. This involves two aspects: understanding given visual marks (\emph{e.g.}, masks or bounding boxes) and identifying whether they are appropriate; reasoning about target visual regions and generating corresponding visual marks. We also observe that existing LMMs seem to perform better on the former task. A straightforward explanation is that verification is more fundamental than exploration~\cite{gandhi2025cognitive}. 

With this in mind, we seek to improve performance by proposing a ``search against verifiers'' pipeline. It involves two roles that collaborate in an iterative process, an \emph{Actor} and a \emph{Verifier}, both built using the same LMM. \emph{Actor} generates affordance predictions, while \emph{Verifier} assesses their correctness. If \emph{Verifier} does not approve the current prediction, it provides feedback to \emph{Actor}, including explanations, suggested affordances, and contextual information. Based on this feedback, \emph{Actor} refines the affordance predictions and the iteration continues. The loop terminates when \emph{Verifier} agrees or the maximum number of refinement steps is reached. Notably, we prompt \emph{Verifier} with visualized affordance predictions rather than the original text-form output from \emph{Actor}, to better leverage the capabilities of LMM.

Overall, the contributions of this work are as follows.
\begin{itemize}
    \item We introduce a new affordance prediction task. It necessitates reasoning about affordances in response to instructions from an egocentric view, which is crucial for intelligent open-vocabulary robotic manipulation.
    \item We present a new dataset comprising fifteen thousand object-instruction-affordance triplets. It is essential for providing insights into model capabilities and facilitating future research. 
    \item We develop a ``search against verifiers'' pipeline based on LMM to generate affordances, which progressively self-refines the predictions. It can serve as a strong baseline for the instruction-oriented affordance prediction task,~de-monstrating superior performance.
\end{itemize}
\section{Related Works}
     
\textbf{Affordance}
has long been considered the inherent properties of objects that suggest possible actions. A branch of work stemming from this definition initially succeeds in a fully supervised manner~\cite{DBLP:conf/cvpr/ChuangL0F18, DBLP:conf/icra/DoN018, DBLP:conf/cvpr/FangWYSL18} with large-scale annotated datasets~\cite{DBLP:conf/cvpr/ChuangL0F18, DBLP:conf/cvpr/DengXW0J21}. Recent research interest also includes weakly-supervised methods~\cite{DBLP:conf/cvpr/0008JSS23}, or one-shot learning~\cite{DBLP:conf/cvpr/0008SSJ24} to reduce annotation costs. An emerging and promising area is the automatic extraction of affordance knowledge through the observation of human interactions with objects in natural environments. HOI~\cite{DBLP:conf/cvpr/LiuTM022} first proposes an automatic pipeline to generate data from egocentric videos~\cite{DBLP:journals/pami/DamenDFFFKMMPPW21, DBLP:conf/cvpr/GraumanWBCFGH0L22, DBLP:conf/cvpr/LiuLJLWSLFWY22}, and defines affordance as the forecast of future interaction. Following this paradigm, VRB~\cite{DBLP:conf/cvpr/BahlMCJP23} collects post-contact motion directions in pixel space. Robo-ABC~\cite{DBLP:conf/eccv/JuHZZJX24} creates an affordance memory for retrieving, and RAM~\cite{DBLP:journals/corr/abs-2407-04689} extends this paradigm by retrieving 3D motion directions. Unlike previous works, we believe that affordance should be instruction-dependent, and we conduct our research accordingly.

\textbf{Large Multimodal Model}
has witnessed rapid advancements recently \cite{gpt4,Gemini}. Early foundational works \cite{Alayrac2022Flamingo,Li2023BLIP2,Liu2023LLaVA} establish the core paradigm, namely visual encoder + cross-modal connector + large language model, which remains the dominant framework today. Recent efforts focus on~improving component integration and scalability (\emph{e.g.}, encoder integration, Mixture-of-Experts) \cite{Cambrian2024,Wang2023CogVLM,Baichuan2023Omni}. Parallel to architectural progress, the quality and diversity of data have emerged as critical performance drivers. High-quality instruction tuning datasets \cite{Li2024LLaVAOne,Zhang2024LLaVA,Infinity-MM} substantially improve model capability and generalization. In this work, we select the advanced open-source model Qwen2.5-VL \cite{qwen25-vl} and the closed-source model GPT-4o \cite{gpt4o} as the backbone. 
\section{Method}

\begin{figure*}
    \centering
    \includegraphics[width=0.85\linewidth]{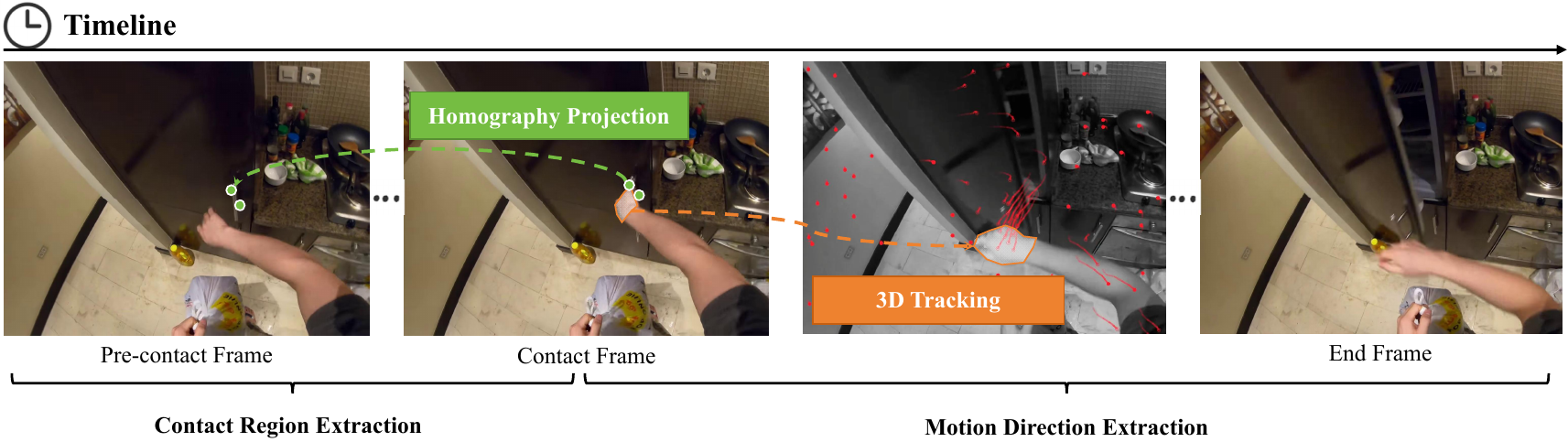}
    \caption{Illustration of the data collection process from egocentric videos. Contact points are projected back to pre-contact frames to identify the contact region without occlusions. 3D tracking is performed to extract the post-contact motion direction.}
    \label{fig:data_collection}
\end{figure*}

\subsection{Basic Principle}
Our primary motivation stems from the observation that affordances should not be treated merely as static properties of objects but rather as dynamic entities closely tied to specific manipulation tasks or instructions. For example, the affordance of a drawer as push-able or pull-able depends on whether the task is to close or open it. Therefore, we introduce a new task alongside a new dataset named EIVA (Egocentric Instruction-oriented Visual Affordance). It necessitates models to predict affordances based on an egocentric image (with complex authentic scenes) and an open-vocabulary instruction. The affordance includes the contact region and the motion direction of manipulation. The affordance annotations in EIVA are generated through an automated pipeline followed by manual reviews, ensuring low-cost and high-quality data for comprehensive analysis.

Recognizing the potential of large multimodal models (LMMs) in handling complex tasks that require rich world knowledge and multimodal understanding, we evaluate both open-source and closed-source LMMs (Qwen2.5-VL~\cite{qwen25-vl} and GPT-4o~\cite{gpt4o}) on the EIVA dataset. Our experiments~reveal that while these LMMs exhibit promising affordance prediction capabilities, they seem to perform better in identifying whether given affordances are appropriate than directly reasoning and generating predictions. Motivated by these observations, we propose a ``search against verifiers'' pipeline, which leverages iterative collaboration between two roles, an \emph{Actor} and a \emph{Verifier}, both constructed using the same LMM. This setup enables progressive self-refinement of affordance predictions through feedback loops.

\subsection{EIVA Dataset}

EIVA consists of 14,861 triplet samples. Each sample contains an open-vocabulary manipulation instruction, an egocentric image (used for extracting the contact region), and affordance annotations encompassing both the contact region and the discretized motion direction. Due to space limitations, the dataset statistics are provided in the Appendix. Please refer to it for more details. 

\subsubsection{Data Preprocessing}

The egocentric videos are obtained from Ego4D~\cite{DBLP:conf/cvpr/GraumanWBCFGH0L22}, Epic-Kitchen~\cite{DBLP:journals/pami/DamenDFFFKMMPPW21} and HOI4D~\cite{DBLP:conf/cvpr/LiuLJLWSLFWY22}. We extract video clips on the basis of the start and end times, covering the entire interaction process. The start time marks the moment when the hand first contacts the object. By convention, we refer to the corresponding frame as the contact frame. In addition, since occlusion of objects by hands frequently occurs, we include the $N$ frames preceding the contact frame within the extracted video clips. In these $N$ frames, typically the hand has not yet made contact with the object, allowing a more accurate capture of the contact regions. We designate them as pre-contact frames.

Pre-processing primarily involves locating the interaction between the hand and the object in extracted video clips, providing essential information for subsequent steps. We first utilize a hand-object detector to obtain the bounding boxes of the interacting hand and object for every frame in the video clip. The bounding box with the highest detection confidence among all frames is selected. SAM2 is applied to generate corresponding segmentation masks across frames, using the selected bounding box as its input prompt.

Moreover, we manually filter the data based on the detection results (\emph{e.g.}, remove the video clip if the detected object category differs from the one in the narration). In the following sections, we describe an automated pipeline to obtain affordance annotations, as shown in Figure~\ref{fig:data_collection}.

\subsubsection{Contact Region Localization}

The contact region seems to be directly derivable from the object segment in the contact frame. But occlusions frequently occur, which necessitates tracing the contact points back to pre-contact frames without occlusion to achieve more accurate localization.

Specifically, contact points are defined as the peripheral points on the hand segment that intersect with the bounding box of the contacted object. A certain number of contact points are sampled within the contact frame and subsequently projected back to pre-contact frames. This process generally involves estimating the homography between consecutive frames~\cite{DBLP:conf/cvpr/BahlMCJP23}, modeling the effect of camera motion. During this estimation, the hand and object regions are masked out to prevent their motion from interfering with the homography computation. Using homography transformation, the contact points can be projected back frame-by-frame until no detectable contact remains (determined by~\cite{DBLP:conf/cvpr/ShanGSF20}). In pre-contact frames, any projected points falling outside the object segment are considered invalid and are filtered out accordingly. Finally, the valid projected contact points are converted into an affordance map in pixel space via Gaussian blurring.

Although all samples are manually verified to guarantee correctness, we provide visualizations of good cases and bad cases of our automatic annotation and analyze the failure modes in the Appendix for deeper insights.

\begin{figure*}
    \centering
    \includegraphics[width=0.75\linewidth]{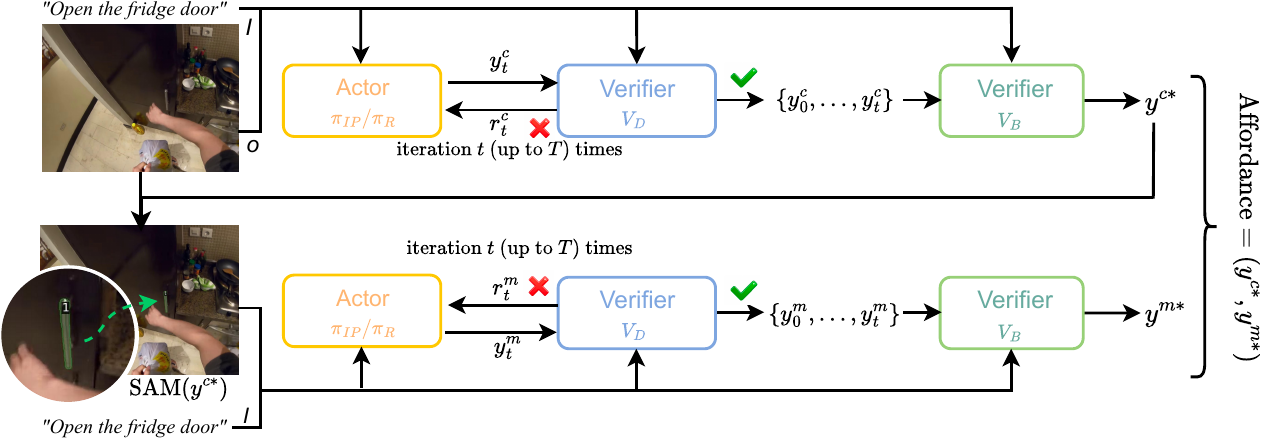}
    \caption{Illustration of the ``search against verifiers'' meta framework, which leverages iterative collaboration between an \emph{Actor} and a \emph{Verifier}. The upper process that searches for the optimal contact region is conducted first. Conditioned on the instruction and the confirmed contact region, the lower process is performed to determine the optimal post-contact motion direction. There might be additional visualization operations, please refer to the sections Coordinate-based and SoM-based for details.}
    \label{fig:search_against_verifiers}
\end{figure*}

\subsubsection{Motion Direction Extraction}
The motion direction after contact is a critical component of the instruction-oriented affordance. For example, the actions of opening and closing a door may point to the same contact region (\emph{e.g.}, the door handle) but require completely opposite motion directions. Unlike previous works that represent motion directions with 2D vectors in pixel space~\cite{DBLP:conf/cvpr/BahlMCJP23}, we employ advanced tools to extract 3D directions, which better aligns with the requirements for applications in embodied systems. To this end, SpatialTracker~\cite{DBLP:conf/cvpr/Xiao0Z0PSZ24} is applied that can lift 2D pixels into 3D and track the motion in the 3D space. We prompt SpatialTracker with the segmentation masks of the hand and object in the contact frame. It generates pixel-level motion trajectories in camera space, \emph{i.e.}, 3D positions of sampled pixels within masks for each frame.

The trajectory cannot directly serve as ground truth because of its limited precision. However, we observe that the overall direction of the trajectory is robust despite noise in positions. Based on this observation, we propose to compute the principal direction of the trajectory to be the motion direction. A Principal Component Analysis (PCA)-like method is introduced. Formally, let $\{p_1, p_2, \dots, p_K\}$ be the $K$ pixels sampled on hands and objects, and $\{\tau(p_1), \tau(p_2),$ $\dots, \tau(p_K)\}$ be their corresponding estimated trajectories. For each trajectory $\tau(p_k)$, we remove outliers using DBSCAN~\cite{DBLP:conf/kdd/EsterKSX96} and truncate it to a maximum length of $10$. Denote the processed trajectory as $\hat{\tau}(p_{k})$, and the covariance matrix of all points in $\hat{\tau}(p_{k})$ as $C$. Eigen-decomposition is then performed on $C$. The eigenvector associated with the largest eigenvalue, denoted as $\dot{\mathbf{v}}_{k} = \arg\max_{\lambda}(\mathbf{v}_{k})$, is designated as the principal direction of $p_k$. The final motion direction $\mathbf{v}$ can be obtained by averaging these principal directions, \emph{i.e.}, $(1 / K)\sum_{i=1}^{K} \dot{\mathbf{v}}_{k}$.

To further mitigate inaccuracies, we discretize the continuous direction space. We define 26 discrete 3D directions as vectors $[x, y, z]$ where $x,y,z \in \{-1,0,1\}$, excluding $[0,0,0]$. These three axes represent forward / backward, downward / upward, and leftward / rightward, respectively. Each continuous 3D direction vector $\mathbf{v}$ is matched to the closest discrete direction with cosine similarity. An example is [backward, upward, leftward].

\subsection{LMM-based Affordance Prediction}

\subsubsection{Meta Framework}

A ``search against verifiers'' framework is developed involving two distinct roles: an \emph{Actor} and a \emph{Verifier}, denoted as $\pi$ and $V$, respectively. They are built using the same LMM and collaborate in an iterative process to refine results. The \emph{Actor} raises affordance proposals and the \emph{Verifier} assesses the correctness and provides feedback. In the following, we present an overview of the framework, and elaborate on two implementations in subsequent sections.

The \emph{Actor} has two types. $\pi_{\text{IP}}$ is designed for generating the \underline{i}nitial affordance \underline{p}roposal at the beginning, and $\pi_{\text{R}}$ for the \underline{r}efinement during the iteration. The \emph{Verifier} also has two types. One is $V_{D}$, which is used for \underline{d}iagnosis, \emph{i.e.}, to verify whether the current affordance prediction is appropriate to complete the instruction in the iteration. The other one, $V_{B}$, is used to select the \underline{b}est prediction from the results of each step at the end of the iteration.

The core idea behind this framework is that scaling test-time compute always benefits performance~\cite{DBLP:journals/corr/abs-2501-19393, DBLP:journals/corr/abs-2411-11504}. The overall process is illustrated in Figure~\ref{fig:search_against_verifiers}. There are two iterative stages, predicting the contact region first and then the motion direction. Formally, denote the visual observation (image) and textual instruction as $o$ and $I$, respectively. Let $T$ be the maximum number of iterations. For \underline{c}ontact region, the \emph{Actor} $\pi_{\text{IP}}$ produces an initial contact region proposal $y_0^{c}$, and the iteration begins. At the $t$-th step ($t \leq T$), the \emph{Verifier} $V_{D}$ accesses the current proposal $y_t^c$. If $y_t^c$ is considered correct, the iteration stops, and the final prediction of the contact region $y^{c*}$ is determined with $V_{B}$. Otherwise, based on the feedback $r_t^c$ from $V_{D}$, the \emph{Actor} $\pi_{R}$ continues to generate a refined proposal $y_{t+1}^c$. The iteration process for the motion direction follows a similar pattern but starts after the contact region iteration concludes. Detailed prompting strategy of our proposed reflection pipeline is presented in the Appendix.

\subsubsection{Coordinate-based} \label{subsection3.3.2}

Several works demonstrate that visualizing text sometimes benefits model performance~\cite{DBLP:journals/corr/abs-2501-07542, DBLP:journals/corr/abs-2310-11441}. This process mirrors human cognition, \emph{i.e.}, human naturally form mental images while perceiving and reasoning~\cite{moulton2009imagining}. We adopt this interesting idea, visualizing the text-form proposals produced by \emph{Actor} and using these visualizations to prompt \emph{Verifier} for more effective feedback.

In the contact region prediction stage, both $\pi_{\text{IP}}(o, I)$ and $\pi_{R}$ generate contact region proposal in the form of textual \underline{coordinates} of a bounding box. In the $t$-th iteration step, the visualization of the region proposal $y_t^c$ is a mask marked with the number `1', as illustrated in Figure~\ref{fig:search_against_verifiers}. It is generated by prompting SAM~\cite{kirillov2023segany} with the bounding box of $y_t^c$. Denote this operation as $\text{SAM}(y_t^c)$. The \emph{Verifier} takes the visualization as input rather than the raw textual coordinates, namely $V_D(o,I,\text{SAM}(y_t^c))$ and $V_{B}(o,I, \{\text{SAM}(y_0^c),$ $ \text{SAM}(y_1^c), \dots, \text{SAM}(y_t^c)\})$. Feedback $r_t^c$ includes: which part of an object that $V_D$ suggests to contact, its appearance and probable position, as well as its relative position to surrounding objects. The refinement of the region proposal is achieved by $\pi_{R}(o,I,y_t^c,r_t^c)$.

Subsequently, we obtain the initial motion direction proposal $y_0^m$ by asking $V_{\text{IP}}(\text{SAM}(y^{c*}),I)$ ``\textit{Contacting with the marked area in the image, which direction of motion is appropriate for $\langle instruction \rangle$}?''. Due to the deficiency of current LMMs in processing 3D inputs~\cite{DBLP:conf/eccv/QiDZGHGYM24}, the visualization technique is not used in this iterative stage. We directly pass the textual description of the direction proposal $y_t^m$ to $V_{D}$ and $V_{B}$.

\subsubsection{SoM-based} \label{subsection3.3.3}
The coordinate-based pipeline requires that the LMM possesses robust object grounding capabilities. We propose an alternative to ease the burden of LMM, \emph{i.e.}, first applying SOTA methods to ground object regions and then asking the LMM to select appropriate grounded regions for manipulation. The specific grounding method we adopt is \underline{SoM}~\cite{DBLP:journals/corr/abs-2310-11441}, which uses off-the-shelf segmentation models (\emph{e.g.}, SAM~\cite{kirillov2023segany}) to partition an image into regions and overlays these regions with marks for identification.

In contact region prediction, the observation (image) $o$ is partitioned into $K$ regions. We refer to these regions as candidates to avoid any confusion with the contact region. We use the visualization method in SoM~\cite{DBLP:journals/corr/abs-2310-11441}, in which each candidate is color-coded and marked with a numerical label. Denote SoM-based visualization as $\text{SoM}(o)$. Both $\pi_{\text{IP}}(\text{SoM}(o), I)$ and $\pi_{R}(\text{SoM}(o), I, y_t^c,r_t^c)$ are asked to select a set of these candidates (based on $I$) to form the proposal $y_t^c$. The \emph{Verifier} has similar formulations: $V_D(\text{SoM}(o),I,y_t^c)$ and $V_{B}(\text{SoM}(o),I, \{y_0^c, y_1^c, \dots, y_t^c\})$.

The visualization of $y^{c*}$ using SoM with mark `1' is identical to $\text{SAM}(y^{c*})$. Consequently, we can reuse the iteration process from the coordinate-based pipeline for the motion direction prediction.
\section{Experiments}

\subsection{Benchmark Setting}
We conduct zero-shot instruction-oriented affordance prediction experiments on the EIVA dataset. Given an egocentric image and a manipulation instruction, the model is required to locate the contact region of object and predict the motion direction (with which the task can be completed). 

We compare the two methods proposed in Section \textit{Coordinate-based}~\ref{subsection3.3.2} and Section \textit{SoM-based}~\ref{subsection3.3.3} (refer to the Appendix for implementation details) against five SOTA end-to-end approaches: VRB~\cite{DBLP:conf/cvpr/BahlMCJP23}, LOCATE~\cite{DBLP:conf/cvpr/0008JSS23}, OOAL~\cite{DBLP:conf/cvpr/0008SSJ24}, 3DOI~\cite{DBLP:conf/iccv/0001F23}, and ManipVQA~\cite{DBLP:conf/iros/0004P00HG0D24}. OOAL does not support open-vocabulary prediction. We map the operations in EIVA to the affordance categories in OOAL, and then conduct the test. To make the raw masks predicted by LMM-based methods compatible with common evaluation metrics, we convert them into affordance maps (contact region heatmap). This is done by performing grid sampling on the predicted masks and then applying Gaussian blur.

To evaluate the location of the contact region, we adopt the same post-processing method as~\cite{DBLP:conf/cvpr/0008SSJ24}, which includes downsampling and normalizing the contact region heatmap to a standard resolution of $(224, 224)$, making sure that it sums up to $1$. 

Four common performance metrics are used: SIM~\cite{DBLP:journals/ijcv/SwainB91}, AUC-Judd (AUC-J)~\cite{DBLP:conf/iccv/JuddEDT09}, Normalized Scanpath Saliency (NSS)~\cite{PETERS20052397} and Cosine Similarity (CS). The first three are used for evaluating the contact region, while the last one assesses the motion direction. Detailed descriptions of these metrics are provided in the Appendix.

\begin{table} \scriptsize
    \centering
    \begin{tabular}{llllc}
        \toprule
        \multirow{3}{*}{\textbf{Method}} & \multicolumn{3}{c}{\textbf{CR}} & \multicolumn{1}{c}{\textbf{MD}} \\
        \cmidrule(lr){2-4} \cmidrule(lr){5-5}
         & SIM$\uparrow$ & NSS$\uparrow$ & AUC-J$\uparrow$ & CS$\uparrow$ \\
        \midrule
         3DOI (RandomQuery@15) & 0.019 & -0.036 & 0.491 & - \\
         OOAL & 0.218 & 1.810 & 0.836 & -  \\
         VRB & 0.188 & 1.008 & 0.681 & -  \\
         LOCATE & 0.241 & 2.216 & 0.933 & -  \\
         ManipVQA & 0.376 & 2.131 & 0.892 & - \\
         \midrule
         Qwen2.5-VL-7B & \underline{0.519} & \underline{2.915} & \underline{0.953} & 0.012  \\
         Qwen2.5-VL-7B-Reflection & \textbf{0.530} & \textbf{2.972} & \textbf{0.957} & 0.018  \\
         GPT4o-SoM & 0.355 & 1.892 & 0.880 & \underline{0.040}  \\
         GPT4o-SoM-Reflection & 0.391 & 2.398 & 0.902 & \textbf{0.060}  \\
         \bottomrule
    \end{tabular}
    \caption{Affordance prediction results on EIVA. CR represents for ``Contact Region'', and MD represents for ``Motion Direction''. $\uparrow$ denotes that larger values of the metric imply better model performance. The \textbf{best} and \underline{second-best} results are highlighted in bold and underlined, respectively.}
    \label{tab:eiva_fullset}
\end{table}

\begin{table*} \small
    \centering
    \begin{tabular}{llllclllc}
        \toprule
        \multirow{3}{*}{\textbf{Method}} & \multicolumn{4}{c}{\textbf{Real-World}} & \multicolumn{4}{c}{\textbf{Laboratory}} \\
        \cmidrule(lr){2-5} \cmidrule(lr){6-9}
        & \multicolumn{3}{c}{\textbf{CR}} & \multicolumn{1}{c}{\textbf{MD}} & \multicolumn{3}{c}{\textbf{CR}} & \multicolumn{1}{c}{\textbf{MD}} \\
        \cmidrule(lr){2-4} \cmidrule(lr){5-5} \cmidrule(lr){6-8} \cmidrule(lr){9-9}
                & SIM$\uparrow$ & NSS$\uparrow$ & AUC-J$\uparrow$ & CS$\uparrow$ & SIM$\uparrow$ & NSS$\uparrow$ & AUC-J$\uparrow$ & CS$\uparrow$ \\
        \midrule
         3DOI (RandomQuery@15) & 0.023 & -0.005 & 0.501 & - & 0.018 & -0.040 & 0.490 & - \\
         VRB  & 0.100 & 0.510 & 0.593 & -  & 0.200 & 1.072 & 0.692 & - \\
         OOAL & 0.126 & 0.845 & 0.742 & -  & 0.237 & 1.824 & 0.851 & - \\
         LOCATE & 0.143 & 1.068 & 0.800 & -  & 0.254 & 2.365 & 0.950 & - \\
         ManipVQA & 0.182 & 1.232 & 0.787 & - & 0.401 & 2.247 & 0.906 & - \\
         \midrule
         Qwen2.5-VL-7B & 0.273 & \underline{1.918} & \underline{0.887} & 0.042  & \underline{0.539} & \underline{2.952} & \underline{0.958} & 0.008 \\
         Qwen2.5-VL-7B-Reflection & \underline{0.285} & \textbf{2.015} & \textbf{0.903} & 0.051  & \textbf{0.549} & \textbf{3.003} & \textbf{0.962} & 0.014 \\
         GPT4o-SoM & 0.282 & 1.623 & 0.820 & \underline{0.200}  & 0.363 & 2.141 & 0.884 & \underline{0.019} \\
         GPT4o-SoM-Reflection & \textbf{0.322} & 1.910 & 0.848 & \textbf{0.235}  & 0.400 & 2.462 & 0.910 & \textbf{0.037} \\
         \bottomrule
    \end{tabular}
    \caption{Affordance prediction results for the Real-World and Laboratory subsets.}
    \label{tab:eiva_subsets}
\end{table*}

\begin{figure*}[t]
    \centering
    \includegraphics[width=0.85\linewidth]{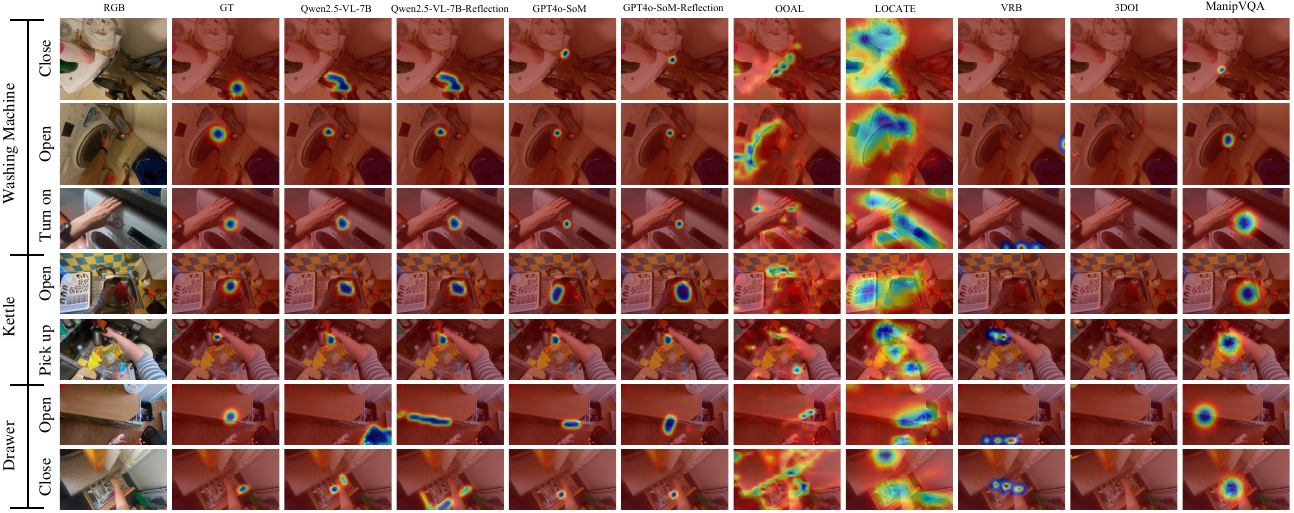}
    \caption{Qualitative comparison. Colder color indicates higher possibility in the affordance map. Each row corresponds to a specific $\langle object, instruction \rangle$ pair, while each column displays the image input or prediction from a particular method.}
    \label{fig:qualitative_comparison}
\end{figure*}

There are actually differences among the videos from the Ego4D, Epic-Kitchens, and HOI4D datasets. The former two datasets feature domestic and real-world environments with substantial visual distractors. Meanwhile, HOI4D consists of laboratory-configured scenes that have relatively fewer distractors, with interactions framed to center on the target objects. Consequently, we split the EIVA dataset into two distinct subsets: 1,704 samples from Ego4D and Epic-Kitchens, termed ``\emph{Real-World}'', and the remaining samples from HOI4D, termed ``\emph{Laboratory}''. We also report performance on these two subsets for comprehensive analysis.

\subsection{Quantitative and Qualitative Comparisons}
\label{sec:quantitative_qualitative}


\begin{figure}[t]
    \centering
    \includegraphics[width=0.95\linewidth]{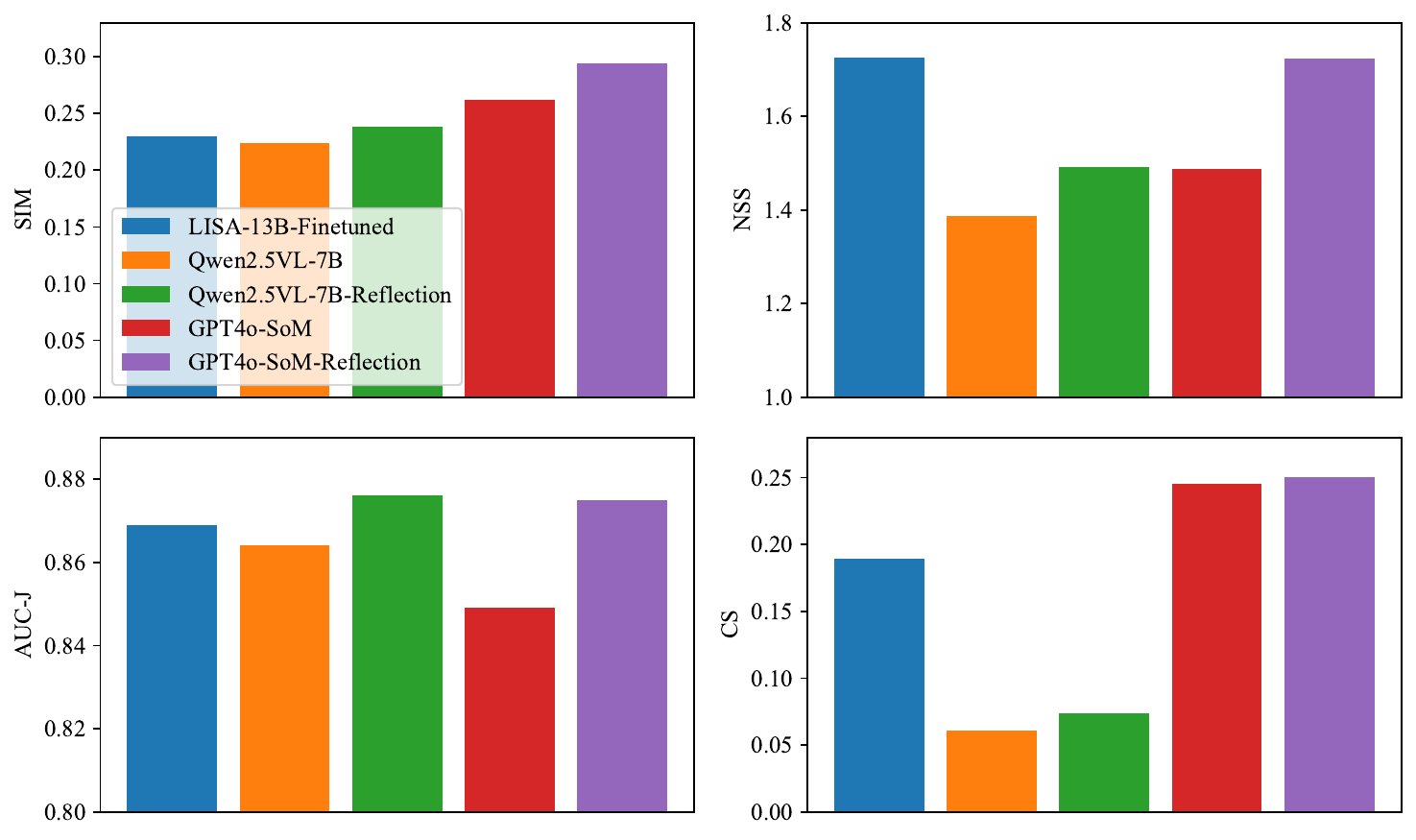}
    \caption{Comparison of affordance prediction results with a fine-tuned specialist on the test set of EIVA.}
    \label{fig:testset}
\end{figure}

The quantitative results are shown in Table~\ref{tab:eiva_fullset}, where GPT4o-SoM and Qwen2.5-VL-7B serve as baselines that directly use an LMM to predict affordances. The entire EIVA dataset is included. From this table four main observations can be obtained. \underline{First}, LMM-based methods are indeed promising affordance predictors, \emph{e.g.}, Qwen2.5-VL-7B always delivers higher results than non-LMM competitors even without any fine-tuning or adaptation. \underline{Second}, the proposed ``search against verifiers'' framework is effective. Qwen2.5-VL-7B-Reflection and GPT4o-SoM-Reflection consistently outperform the baseline methods in terms of performance. \underline{Third}, motion direction prediction is more difficult than contact region localization for LMM-based methods (we omit competitor results as they fail to produce 3D orientations). The results are considered acceptable, as the CS scores are consistently positive. This indicates that the predicted motion directions align with the ground truth rather than being opposite, demonstrating their practical utility. \underline{Last but not the} \underline{least}, Qwen2.5-VL performs better on the first task, while GPT-4o excels in the other. We believe that these results can provide some new insights into the capabilities and limitations of current LMMs.

The experimental results on the two subsets of EIVA are listed in Table~\ref{tab:eiva_subsets}. The observations and conclusions derived from Table~\ref{tab:eiva_fullset} still hold in Table ~\ref{tab:eiva_subsets}. Moreover, it can be observed that the improvement brought by LMM is more significant on the harder ``Real-World'' subset. It also demonstrates the LMM's generalization ability in handling complex real-world scenarios, attributed to the extensive world knowledge and robust model capabilities.

In Figure~\ref{fig:qualitative_comparison}, we qualitatively analyze the contact regions predicted by different models. Three objects with different operations are included (refer to the Appendix for more qualitative comparison). The results show that LMM-based models generally predict more compact contact areas compared to their competitors. Among them, Qwen2.5-VL-7B-Reflection demonstrates superior performance. For example, it accurately identifies the contact regions for actions such as opening or picking up a kettle, locating the top area and the handle, respectively. The model even knows that you can open or close the drawer by interacting with any part of its edge.


One may wonder whether performance can be further improved by training on EIVA. To answer this question, we divide the entire EIVA dataset into two parts: eight object categories (lid, blender, washing, box, ladder, button, scale, and dishwasher) from the ``Real-World'' subset are allocated for testing, while the rest serve as the training set. The categories of these two sets are non-overlapped. LISA~\cite{DBLP:conf/cvpr/LaiTCLY0J24}, a powerful LLM-enhanced segmentation model, is selected as the backbone. 
Details of fine-tuning LISA-13B are presented in the Appendix. We compare Qwen2.5-VL-7B-Reflection and GPT4o-SoM-Reflection with the fine-tuned LISA-13B on the test set. Comparisons are shown in Figure~\ref{fig:testset}. LISA-13B-Finetuned achieves competitive performance, but still inferior to GPT4o-SoM-Reflection. This might be because the training data is not sufficiently large.

\subsection{Comparisons on Existing Datasets}
We conduct experiments to evaluate our approach in a zero-shot way on existing affordance prediction datasets. Specifically, Table~\ref{tab:agd20k} lists the comparisons on the AGD20K dataset~\cite{DBLP:conf/cvpr/LuoZZ0T22}. We do not use the KLD (Kullback-Leibler Divergence) metric~\cite{PETERS20052397} as it is known to be sensitive to the tails of distributions~\cite{ben2015kullback}. Our LMM-based affordance predictor still achieves SOTA performance, particularly excelling in unseen categories where competing methods often struggle with generalization. It reaches the best performance in SIM, and the second best in NSS. This further demonstrates the effectiveness and generalization of our method.

\begin{table} \scriptsize
    \centering
    \begin{tabular}{lllll}
        \toprule
        \multirow{2}{*}{\textbf{Method}} & \multicolumn{2}{c}{\textbf{Seen}} & \multicolumn{2}{c}{\textbf{Unseen}} \\
        \cmidrule(lr){2-3} \cmidrule(lr){4-5}
                & SIM$\uparrow$ & NSS$\uparrow$ & SIM$\uparrow$ & NSS$\uparrow$ \\
        \midrule
         Hotspots & 0.278 & 0.615 & 0.237 & 0.577 \\
         Cross-view-AG & 0.334 & 0.927 & 0.285 & 0.829 \\
         Cross-view-AG+ & 0.342 & 0.981 & 0.279 & 0.882 \\
         MaskCLIP & 0.169 & 0.041 & 0.152 & 0.047 \\
         SAN & 0.357 & 0.941 & 0.351 & 1.022 \\
         ZegCLIP & 0.387 & 1.001 & 0.361 & 1.042 \\
         OOAL & \textbf{0.577} & \textbf{1.745} & 0.461 & \textbf{1.503} \\
         \midrule
         Qwen2.5-VL-7B              & 0.423 & 1.209 & 0.461 & 1.333 \\
         Qwen2.5-VL-7B-Reflection   & 0.469 & \underline{1.343} & \textbf{0.481} & 1.416 \\
         GPT4o-SoM                  & 0.459 & 1.253 & 0.462 & 1.408 \\
         GPT4o-SoM-Reflection       & \underline{0.470} & 1.296 & \underline{0.480} & \underline{1.475}\\
         \bottomrule
    \end{tabular}
    \caption{Comparison with state of the art on AGD20K dataset.}
    \label{tab:agd20k}
\end{table}

\subsection{Performance in Robot Simulation Trails}

\begin{table}[t]
\scriptsize
\centering
\begin{tabular}{lccccc}
\toprule
\textbf{Method} & 
\includegraphics[width=1.5em]{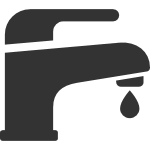} &
\includegraphics[width=1.5em]{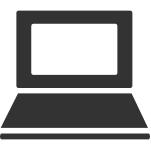} &
\includegraphics[width=1.5em]{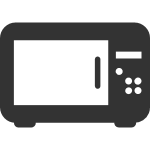} &
\includegraphics[width=1.5em]{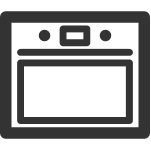} &
\includegraphics[width=1.5em]{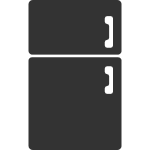} \\
\midrule
Where2Act & 0.17 & 0.17 & 0.00 & 0.00 & 0.17 \\
ManipLLM & 0.17 & 0.33 & 0.50 & 0.33 & 0.67 \\
Qwen2.5VL-7B & 0.50 & 0.83 & 0.17 & 0.33 & 1.00 \\
Qwen2.5VL-7B w/ Reflection & 0.67 & 1.00 & 0.50 & 0.33 & 1.00 \\
GPT4o-SoM & 0.50 & 0.50 & 0.00 & 0.00 & 0.67 \\
GPT4o-SoM w/ Reflection & 0.50 & 0.67 & 0.00 & 0.33 & 0.83 \\
\midrule
\textbf{Method} &
\includegraphics[width=1.5em]{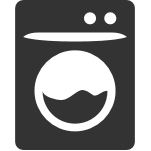} &
\includegraphics[width=1.5em]{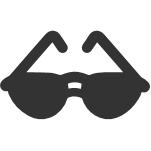} &
\includegraphics[width=1.5em]{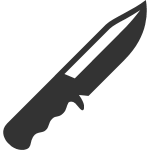} &
\includegraphics[width=1.5em]{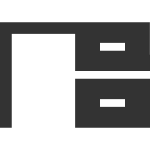} &
\includegraphics[width=1.5em]{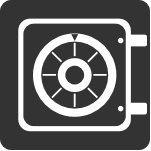} \\
\midrule
Where2Act & 0.17 & 0.00 & 0.00 & 0.33 & 0.17 \\
ManipLLM & 0.67 & 0.17 & 0.00 & 0.67 & 0.67 \\
Qwen2.5VL-7B & 1.00 & 0.00 & 0.33 & 0.83 & 1.00 \\
Qwen2.5VL-7B w/ Reflection & 1.00 & 0.17 & 0.50 & 1.00 & 1.00 \\
GPT4o-SoM & 0.67 & 0.50 & 0.50 & 0.33 & 0.83 \\
GPT4o-SoM w/ Reflection & 0.67 & 0.82 & 0.78 & 0.50 & 0.83 \\
\bottomrule
\end{tabular}
\caption{Success rates of methods across different objects.}
\label{tab:robo_success_rate}
\end{table}

To further validate performance of our proposed pipeline, we conduct robot simulation experiments in SAPIEN following the setup of ManipLLM~\cite{DBLP:conf/cvpr/0020ZGGL0ZLD24}.
In each trial, the VLMs are provided with an RGB-D observation of the scene and are asked to predict contact area and post-contact motion to finish the task described in natural language.
To obtain the precise end-effector contact point and pose, we ground the predicted contact area using GraspNet~\cite{DBLP:journals/trob/FangWFGLYLXL23}.
For detailed evaluation protocol, please refer to the Appendix.


The overall success rate is presented in Table~\ref{tab:robo_success_rate}.
VLMs outperforms the baselines even without the reflection process.
This is largely due to their ability of understanding and following instructions.
In contrast, the baselines can only ``manipulate'' the object with on purpose.
We also observe a consistent lift on the performance of VLMs by applying our proposed reflection pipeline.
Qualitative visualizations of how reflection corrects the wrong propose can be found in the Appendix.

\section{Conclusion}
In this work, we have introduced a new affordance prediction task. This task is more challenging than the vanilla affordance prediction task, as it requires the model to reason about action possibilities according to specific manipulation instruction. To facilitate effective evaluation, we have presented a new dataset for this task. An automated pipeline is developed to extract affordances, thereby reducing annotation costs. Furthermore, we have proposed an LMM-based affordance prediction pipeline comprised of an \emph{Actor} and a \emph{Verifier}, both constructed using the same LMM. \emph{Actor} explores potential affordances, while \emph{Verifier} evaluates their correctness and provides feedback. This iterative collaboration between the two components progressively refines the affordance predictions. Our approach establishes a robust baseline for the introduced instruction-oriented affordance prediction task. We believe that this work can give useful insights to the community.

\bibliography{aaai2026}

\clearpage
\appendix

\section{Justification of Our New Definition of Task-Oriented Affordance}

While the classical Gibsonian view defines affordance as the action possibilities the environment offers to an agent \textit{regardless of the agent’s intentions}, this definition, though foundational, does not provide sufficient operational guidance for robotic behavior in goal-conditioned tasks.

In the context of robotics, simply knowing that a door handle affords ``grasping'' does not specify \textit{how} to perform a meaningful action such as ``open the door.''
That is, affordance as a static property of the object lacks the specificity required for task-oriented interaction.

To address this, recent research has proposed ``agent-centric affordances''~\cite{DBLP:conf/iccv/MoGM0T21, DBLP:conf/cvpr/0020ZGGL0ZLD24}, which define affordances as all possible point-direction pairs on an object surface that can cause part motion when acted upon.
These works consider an interaction successful if the manipulated part exhibits motion in the intended direction.
While this formulation is closer to ours, such approaches cannot represent affordances in a \textit{task-conditioned} manner.
Furthermore, these methods often rely on large-scale sampling in simulation to generate training data, limiting their applicability to real-world settings.

In contrast, our proposed \textbf{task-oriented affordance} formulation builds upon the agent-centric perspective, extending it in two important directions:
\begin{enumerate}
    \item \textbf{Language-conditioned affordances}: By conditioning affordance predictions on natural language task descriptions, our method provides actionable and task-specific guidance, enabling robots to choose different contact regions and manipulation strategies for different goals (e.g., “close the fridge” vs. “open the fridge”).
    \item \textbf{Real-world generalization}: Our model predicts affordances directly from real-world visual observations and leverages internet-scale language-vision data. This allows for broader applicability beyond simulation and significantly improves transfer to real-world manipulation scenarios.
\end{enumerate}

Moreover, RT-Affordance~\cite{DBLP:journals/corr/abs-2411-02704} has proposed defining affordance as the pose of the robot's end-effector at key stages of a task.
While this definition shares our task-oriented motivation, the representation differs—ours captures contact-area and post-contact motion, whereas theirs focuses on key-frame poses.

In summary, while our task-oriented definition of affordance departs from the classical, goal-agnostic view, it is a necessary and increasingly adopted shift in the robotics manipulation community.
It enables more precise, goal-driven control policies and better aligns with the practical requirements of embodied agents.

\section{Robot Simulation Trails}

\begin{figure}[t]
    \centering
    \includegraphics[width=1\linewidth]{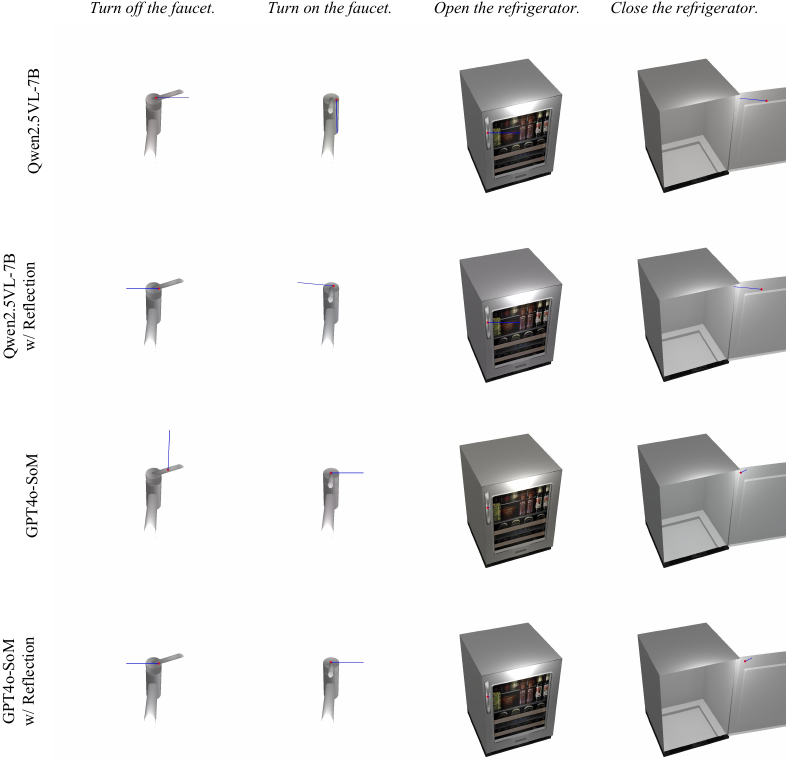}
    \caption{Visualization of the predictions in robot simulation trails.}
    \label{fig:robo_sim}
\end{figure}

Unlike ManipLLM, which considers any motion (either translational or rotational) of movable parts as a successful manipulation, our evaluation pipeline is specifically designed to reflect \textbf{task-oriented success}.
That is, we extend the original setup by assigning distinct tasks to the same object and viewpoint, and manually annotate the object’s links relevant to each specific task.
During evaluation, only motion of these task-relevant links is considered a success.
Manipulation that moves irrelevant parts is not rewarded.
We select ten object categories and sample three object instances from each category.
Each object instance is evaluated on two different tasks (e.g, open / close the fridge door), performed from the same fixed camera viewpoint, to assess the model’s ability to disambiguate affordances based on task intent.

The qualitative examples shown in Figure~\ref{fig:robo_sim} demonstrate that VLMs are capable of predicting appropriate task-conditioned affordances from the same viewpoint.
For instance, the model successfully infers that opening a fridge requires pulling the handle, whereas closing it involves pulling on the door edge when no handle is available.
Furthermore, when manipulating the faucet, our proposed pipeline improves the accuracy of post-contact motion prediction.

\section{Implementation details}

We instantiate our proposed pipeline in Section~\ref{subsection3.3.2} using Qwen2.5-VL-7B~\cite{qwen25-vl} as both the \emph{Actor} and the \emph{Verifier}. We name this instantiation Qwen2.5-VL-7B-Reflection. In contrast, we note that GPT-4o lacks sufficient grounding capability on the EIVA images. Therefore, we chose to pair the SoM-based variant (described in Section~\ref{subsection3.3.3}) with GPT-4o~\cite{gpt4o}. This is referred to as GPT4o-SoM-Reflection. For both Qwen2.5-VL-7B-Reflection and GPT4o-SoM-Reflection, the maximum number of iterations, \emph{i.e.}, $T$, is set as $3$. ``Numeric'' marks, colored masks, and mask boundaries are utilized for visualization (visual prompting on the input image). For GPT4o-SoM-Reflection, we opt for SAM for image partition, as it provides the finest granularity among all candidates~\cite{DBLP:journals/corr/abs-2310-11441}.

\section{Additional Qualitative Results on EIVA15k}

\begin{figure*}[t]
    \centering
    \includegraphics[width=1\linewidth]{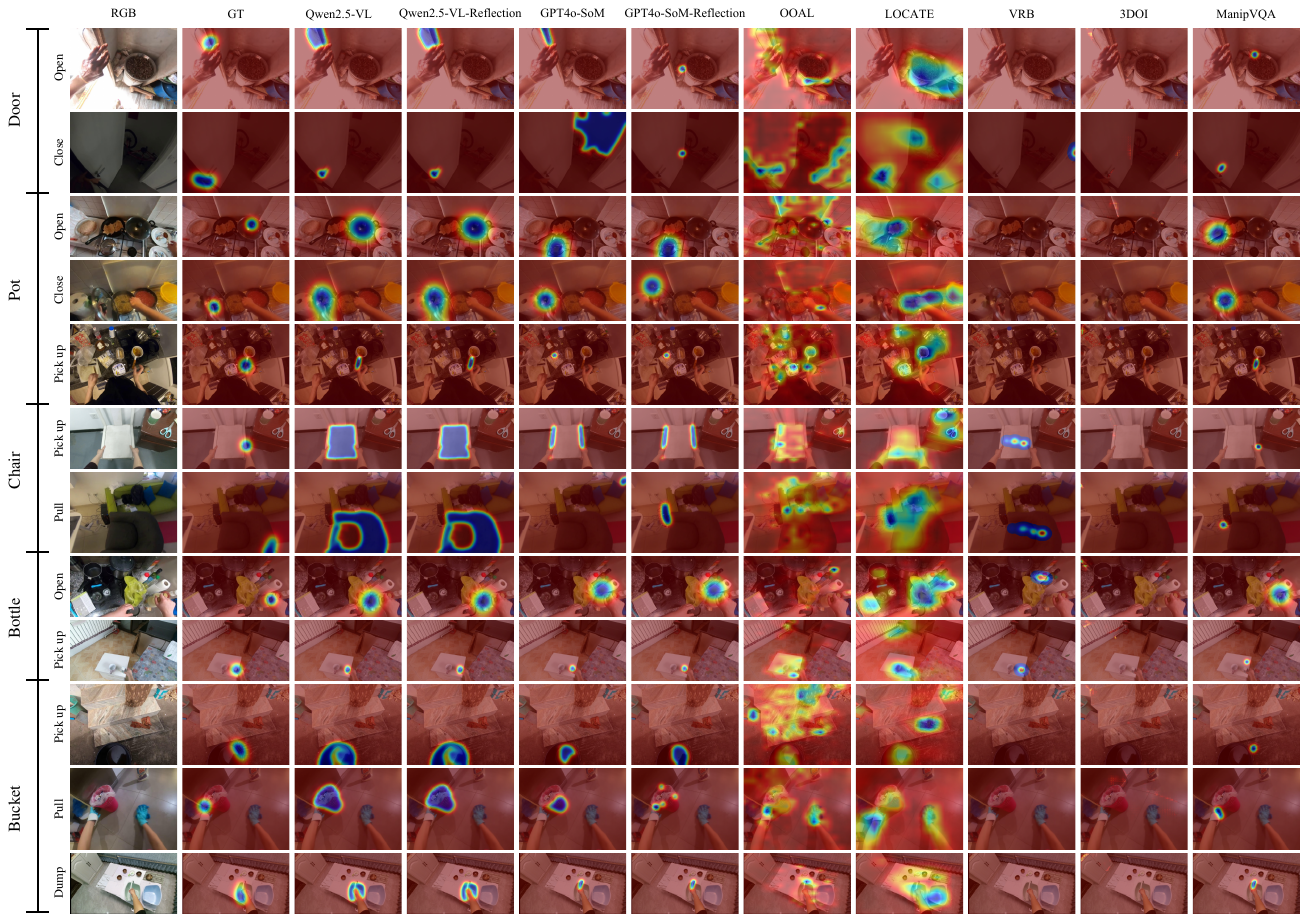}
    \caption{Additional qualitative comparison. Each row corresponds to a specific $\langle object, instruction \rangle$ pair, while each column displays the image input, ground truth (GT), or prediction from a particular method.}
    \label{fig:additional_qualitative}
\end{figure*}

Qualitative comparison on additional $\langle object, instruction \rangle$ pairs in EIVA15k are presented in Figure~\ref{fig:additional_qualitative}.
The results reflects consistent conclusion as we analysed in Section~\ref{sec:quantitative_qualitative}.

\section{Correctness of Automatic Annotation}

\begin{figure*}[t]
    \centering
    \includegraphics[width=1\linewidth]{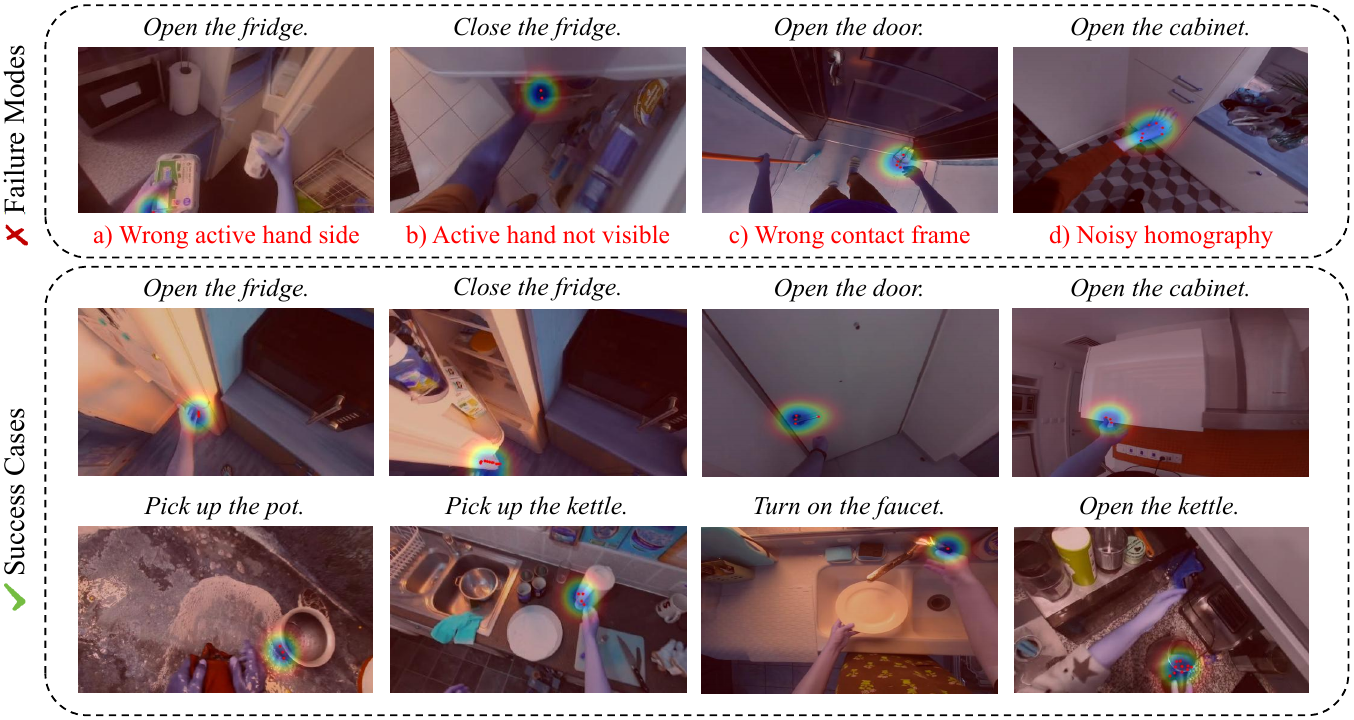}
    \caption{Good cases and failure modes of the automatic annotation pipeline.}
    \label{fig:annotation_cases}
\end{figure*}

Although the correctness of annotations in the EIVA15k is guaranteed by manual verification, we would like to include visualizations of both good cases and failure modes of our proposed automatic annotation pipeline in Figure~\ref{fig:annotation_cases} to provide transparency that help readers assess the annotation reliability.
Since implementation details and theoretical feasibility are fully discussed in Section~\ref{subsection3.3.2}, we mainly discuss the failure modes in this section.

We identified four major failure modes in the automatic annotations.
First, according to the assumption of single-arm operation, we annotate the contact and post-contact motion by tracing the right hand of the camera-wearer.
When the HOI detector~\cite{DBLP:conf/cvpr/ShanGSF20} fails on judging the hand side correctly, our pipeline will trace the wrong hand and thus results in wrong annotations.
As shown in Figure~\ref{fig:annotation_cases}a, our pipeline mistakenly annotated the contact area of base on the left hand of the camera-wearer, and thus the annotated affordance is mismatched with the given task.

Next, because of the narrow horizons of the egocentric point-of-view and occlusions, the active hand, which relates to the given task, may not be visible in the observation.
This leads to failures on HOI detection.
For example, in Figure~\ref{fig:annotation_cases}b the camera-wearer is closing the fridge door but the active hand is occluded by the door itself.
We believe such cases are inevitable due to the fact that visual feedback is not necessary for humans to finish the tasks they are familiar with.

The third failure mode comes with the noisy annotation of the contact frame in upstream datasets (e.g., EpicKitchens).
Our automatic annotation pipeline relays on the assumption that the contact frame is the first frame of HOI.
If the HOI is not yet happened in the ``contact frame'' (e.g., Figure~\ref{fig:annotation_cases}c), the pipeline will regard unreasonable areas as the contact area, and results in wrong annotations.

Lastly, due to the limited precision of the homography projection estimation, the projected contact area can be inaccurate, as demonstrated in Figure~\ref{fig:annotation_cases}d.
Such limited precision stems from the fundamental challenge in feature point matching problem.
That is, 1) the feature points are sparse when the observation is full of pure color; 2) the feature points can hardly match when the camera motion is dramatic between adjacent frames.
These challenges are more common and tricky under egocentric views than the fixed, third person views, which is heavily explored in the computer vision community.
We will improve our pipeline in the future on either reduce the dependency on precise homography estimation or overcome the limitation of the existing feature point matching approaches under egocentric view.

\begin{figure*}[t]
    \centering
    \includegraphics[width=0.8\linewidth]{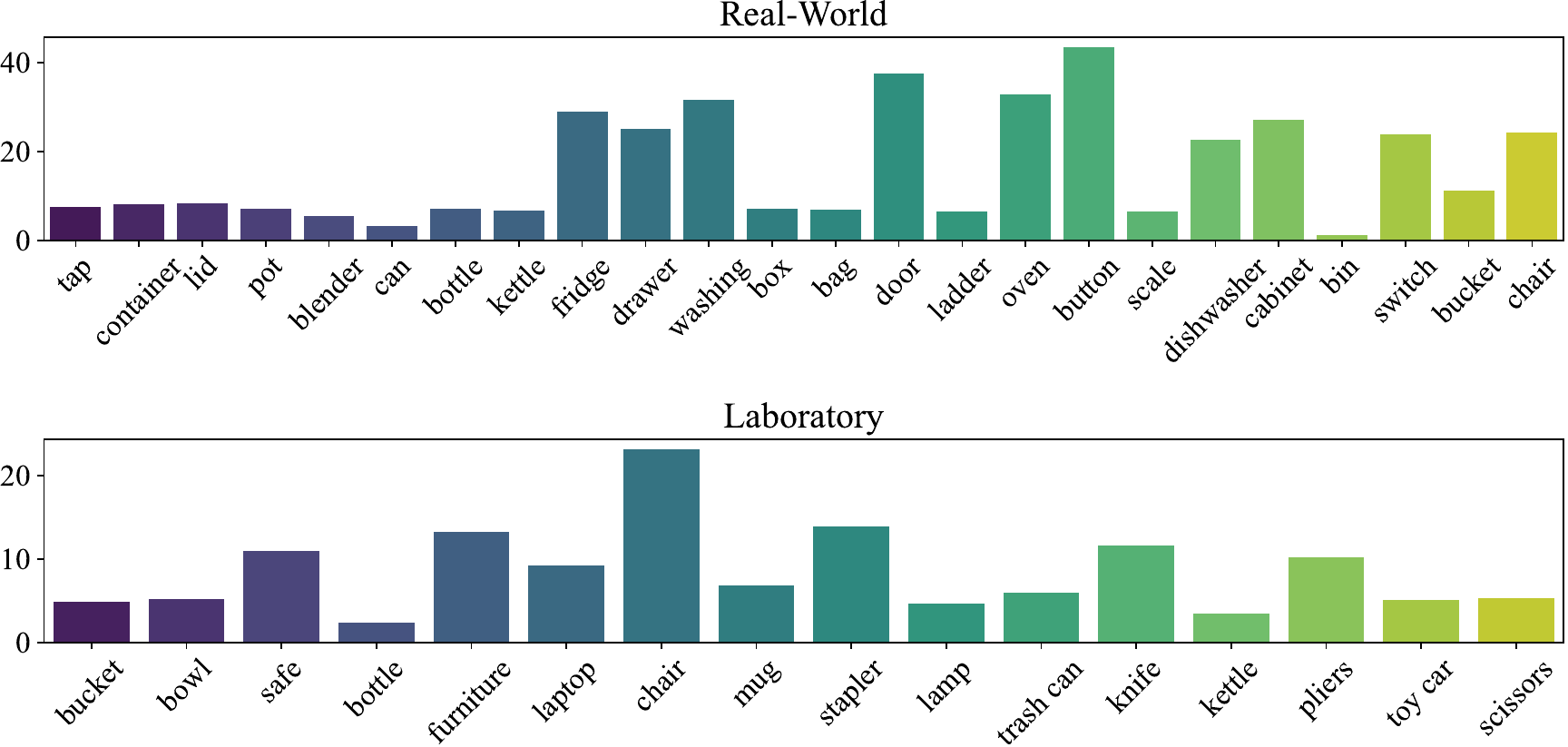}
    \caption{Average image coverage by object category in Real-World and Laboratory subsets.}
    \label{fig:coverage}
\end{figure*}

\begin{figure}[t]
    \centering
    \includegraphics[width=1\linewidth]{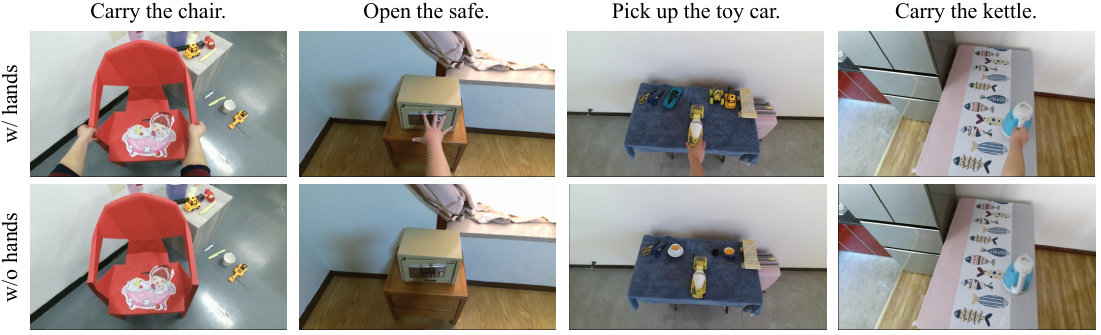}
    \caption{Cases of samples before and after eliminating hands.}
    \label{fig:inpainting_hands}
\end{figure}

\begin{figure}[h]
    \centering
    \includegraphics[width=1\linewidth]{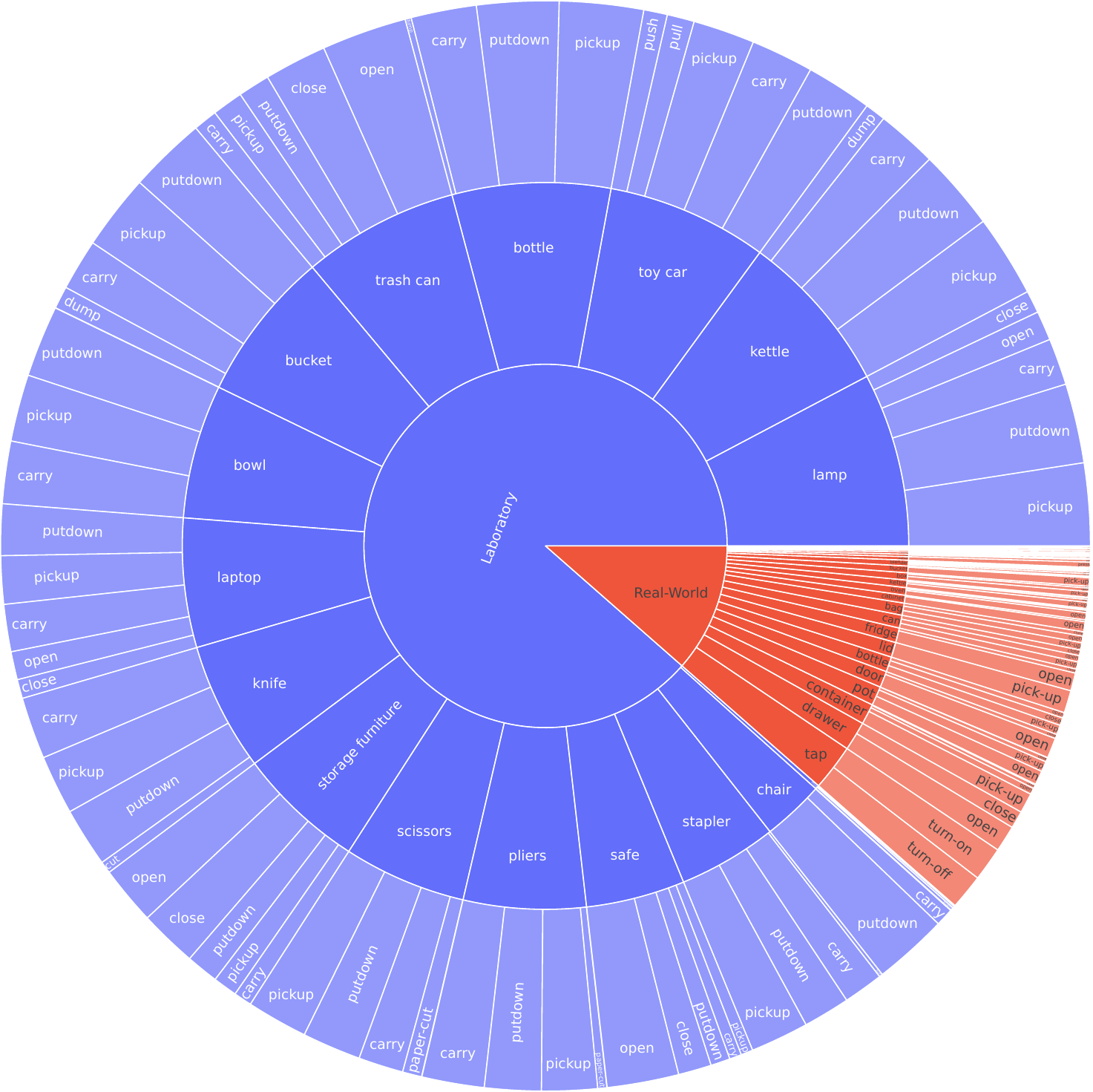}
    \caption{Hierarchical distribution of our EIVA15k dataset, showing subsets (color), object categories (middle ring), specific actions (inner ring), and their frequencies (area size).}
    \label{fig:eiva15k}
\end{figure}

\begin{table}[t]
    \centering
    \begin{tabular}{l|l}
    \toprule
        Resolution & Count \\
        \midrule
        $456 \times 256$ & 1240 \\
        $1440 \times 1080$ & 215 \\
        $1920 \times 1080$ & 23202 \\
        $1920 \times 1440$ & 202 \\
        $2560 \times 1440$ & 2 \\
    \bottomrule
    \end{tabular}
    \caption{Statistic of input resolution in EIVA15k.}
    \label{tab:resolution}
\end{table}

\section{Details of Prompting Strategy}

\begin{figure*}[t]
    \centering
    \begin{subfigure}{1\textwidth}
        \includegraphics[width=1\linewidth]{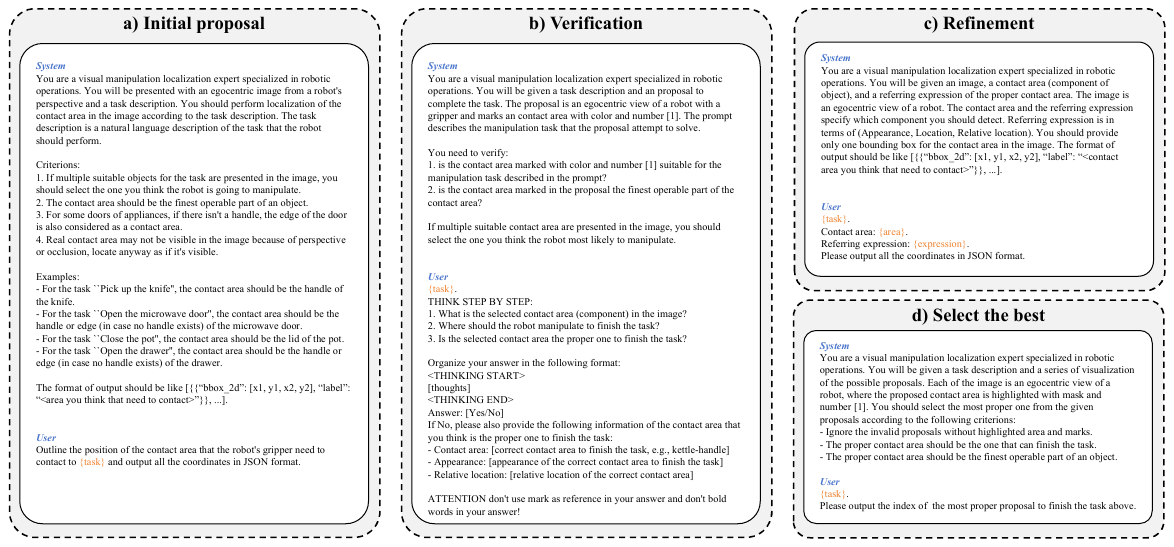}
        \caption{Contact area prediction.}
        \label{fig:prompt-contact}
    \end{subfigure} \\
    \centering
    \begin{subfigure}{1\textwidth}
        \centering
        \includegraphics[width=1\linewidth]{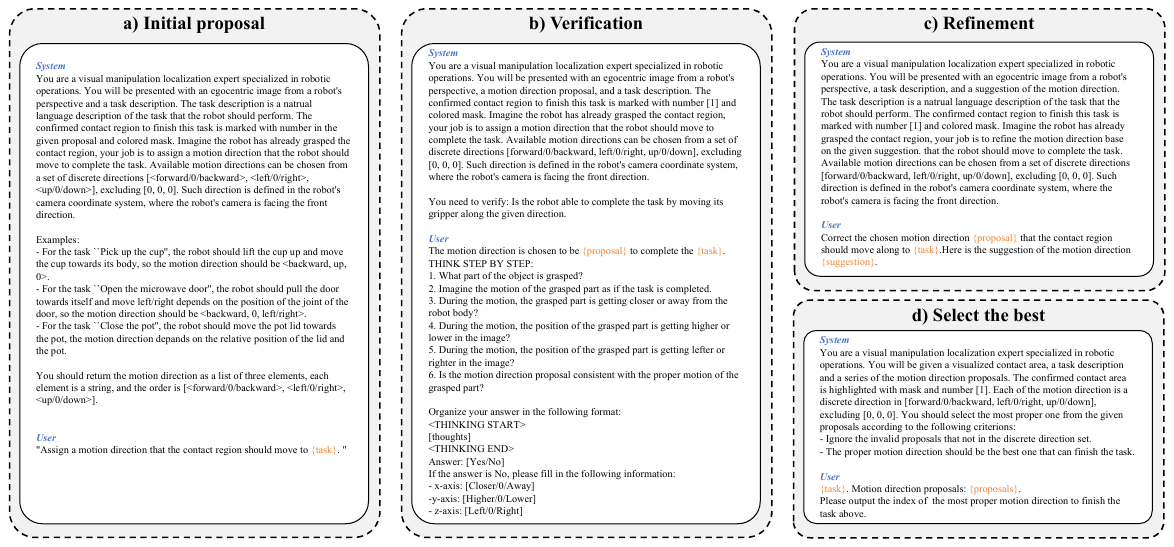}
        \caption{Motion direction prediction.}
        \label{fig:prompt-motion}
    \end{subfigure}
    \caption{Prompting strategy of each step of our affordance prediction pipeline.}
    \label{fig:prompt}
\end{figure*}

To facilitate reproducibility of our work, we here present our contact area prompting strategy in Figure~\ref{fig:prompt-contact} and motion direction prompting strategy in Figure~\ref{fig:prompt-motion}.
Note that these strategies are specific to Qwen2.5-VL.
GPT4o requires some minor adjustments to fit its SoM indexing style of prompting.

\section{Data Source for EIVA}
The egocentric videos are obtained from Ego4D~\cite{DBLP:conf/cvpr/GraumanWBCFGH0L22}, Epic-Kitchen~\cite{DBLP:journals/pami/DamenDFFFKMMPPW21} and HOI4D~\cite{DBLP:conf/cvpr/LiuLJLWSLFWY22}. We selected 22 types of objects from Ego4D, 12 types of objects from Epic-Kitchen, and all 16 types of objects from HOI4D, along with their corresponding videos.
These videos are coupled with timestamped narrations that describe the actions and their respective start and end times. There are two types of actions: hand-object interaction (HOI) and object-object interaction (OOI). Actions involving OOI always include HOI, as a human must hold one object to interact with another. This scenario is particularly difficult (combining OOI and HOI, \emph{e.g.}, using tools to perform tasks), especially within the context of our setup. We leave it for future work. Only HOI videos are included in this work. 

\section{Statistics of EIVA15k}

In this section, we show more statistical information of our EIVA15k.
Figure~\ref{fig:eiva15k} illustrates the proportion of the Real-World and Laboratory subset, proportion of each object category and proportion of possible manipulation that can be applied on the object.
The Laboratory subset forms the major part of EIVA15k.
Each of the object have around four different related manipulation tasks in average, which highlights the concept of task-oriented affordance.

Next, as shown in Figure~\ref{fig:coverage}, most objects in EIVA15k only occupies around ten percent of the image.
This poses challenge of fine-grained visual grounding.
Some other objects cover more in the image since they are large compare to a person.
For example, our eyes will be filled up with the door when we open it.
However, this doesn't means that these samples are easy cases.
The valid contact area, such as handles and edges of the door, still requires precise understanding of the manipulation and visual grounding.

Finally, image resolutions are presented in Table~\ref{tab:resolution}.
Most of our samples have a higher or equal resolution than $1920 \times 1080$.
The high resolution inputs facilitate future evaluation on LMMs which are increasingly stronger at leveraging the richer information in high resolution images.

\section{Ablation Study on Proposal Visualization}

\begin{table}[t]
\scriptsize
\centering
\caption{Ablation study on proposal visualization of Qwen2.5VL-7B w/ Reflection.}
\label{tab:vis_ablation}
\begin{tabular}{llcccc}
\toprule
\multirow{2}{*}{\textbf{Subset}} & \multirow{2}{*}{\textbf{Method}} & \multicolumn{3}{c}{\textbf{CR}} & \textbf{MD} \\
\cline{3-6}
& & \textbf{SIM↑} & \textbf{NSS↑} & \textbf{AUC-J↑} & \textbf{CS↑} \\
\midrule
\multirow{3}{*}{\textbf{Real-World}}
    & Qwen2.5VL-7B & 0.273 & 1.918 & 0.887 & 0.042 \\
    \cline{2-6}
    & \makecell[l]{Qwen2.5VL-7B \\ w/ Reflection} & 0.285 & 2.015 & 0.903 & 0.051 \\
    \cline{2-6}
    & \makecell[l]{Qwen2.5VL-7B \\ w/ Reflection \\ w/o Visualization} & 0.239 & 1.774 & 0.893 & 0.048 \\
\midrule
\multirow{3}{*}{\textbf{Laboratory}}
    & Qwen2.5VL-7B & 0.539 & 2.952 & 0.958 & 0.008 \\
    \cline{2-6}
    & \makecell[l]{Qwen2.5VL-7B \\ w/ Reflection} & \textbf{0.549} & \textbf{3.003} & 0.962 & 0.014 \\
    \cline{2-6}
    & \makecell[l]{Qwen2.5VL-7B \\ w/ Reflection \\ w/o Visualization} & 0.545 & 3.002 & 0.962 & 0.014 \\
\bottomrule
\end{tabular}
\end{table}

One of our major difference to existing agent reflection frameworks is that we visualize the proposal following SoM.
This results in a unique multi-modal reflection framework which is particularly useful in the affordance prediction task, where fine-grained visual grounding ability is neccessary.
To evaluate the improvement made by proposal visualizing, we eliminate the visualizer and compare it with the one with a visualizer.
Note that the ablation can only be applied to Qwen2.5-VL-7B w/ Reflection, since GPT4o-SoM always needs the fully segmented reference image to work.
Table~\ref{tab:vis_ablation} demonstrate that our reflection pipeline consistently reaches better performance on both subsets of EIVA15k.
Moreover, the lack of visualization even cause the decline of the performance in the more challenging Real-World subset.
This indicates that the more complex that visual input is, the more critical the visualizer is for the reflection.

\section{Does VLMs exploit hands as shortcuts?}

\begin{table}[t]
\scriptsize
\centering
\caption{Comparison on EIVA15k subset w/ and w/o hands.}
\label{tab:hand_inpainting_results}
\begin{tabular}{llcccc}
\toprule
\multirow{2}{*}{\textbf{Subset}} & \multirow{2}{*}{\textbf{Method}} & \multicolumn{3}{c}{\textbf{CR}} & \textbf{MD} \\
\cline{3-6}
& & \textbf{SIM↑} & \textbf{NSS↑} & \textbf{AUC-J↑} & \textbf{CS↑} \\
\midrule
\multirow{4}{*}{\textbf{w/ Hands}} 
    & Qwen2.5VL-7B & 0.465 & 2.652 & 0.936 & 0.059 \\
    \cline{2-6}
    & \makecell[l]{Qwen2.5VL-7B \\ w/ Reflection} & 0.469 & 2.664 & 0.937 & 0.069 \\
    \cline{2-6}
    & GPT4o-SoM & 0.306 & 1.992 & 0.874 & 0.063 \\
    \cline{2-6}
    & \makecell[l]{GPT4o-SoM \\ w/ Reflection} & 0.348 & 2.331 & \textbf{0.944} & 0.115 \\
\midrule
\multirow{4}{*}{\textbf{w/o Hands}}
    & Qwen2.5VL-7B & 0.447 & 2.573 & 0.928 & 0.042 \\
    \cline{2-6}
    & \makecell[l]{Qwen2.5VL-7B \\ w/ Reflection} & \textbf{0.470} & \textbf{2.735} & 0.941 & 0.056 \\
    \cline{2-6}
    & GPT4o-SoM & 0.345 & 2.336 & 0.878 & 0.070 \\
    \cline{2-6}
    & \makecell[l]{GPT4o-SoM \\ w/ Reflection} & 0.366 & 2.488 & 0.941 & \textbf{0.151} \\
\bottomrule
\end{tabular}
\end{table}

To assess whether the presence of hands biases the model’s predictions, we constructed a subset of the dataset by randomly sampling 5 examples from each object category (90 samples in total) and performing hand removal via image inpainting using Stable Diffusion~\cite{DBLP:conf/cvpr/RombachBLEO22}.
Examples of the inpainted results are shown in Figure~\ref{fig:inpainting_hands}.

As reported in Table~\ref{tab:hand_inpainting_results}, the VLMs exhibit no performance drop on the inpainted subset; in fact, they achieve slightly better performance compared to the full dataset.
The performance improvement after hand removal may be due to the elimination of visual occlusions caused by the human hand during interaction.
In the original videos, parts of the object can be occluded by the hand, which leads to predicted masks that are spatially shifted or partially incomplete relative to the ground-truth labels.
Once the hand is removed, the object becomes fully visible, and the model can produce a more accurate mask that better aligns with the annotated contact area, thus improving evaluation metrics.

This suggests that the models are not relying on the visual presence of hands as a shortcut for affordance prediction, but are instead leveraging meaningful associations between object appearance and task description.

\section{Details of Finetuning LISA}

\begin{table}[t]
    \centering
    \begin{tabular}{ll}
    \toprule
         Train  & \makecell[l]{tap, lid, container, pot, can, bottle, \\ kettle, fridge, drawer, bag, door, oven, \\ cabinet, bucket, washing, lamp, bowl, \\ laptop, knife, storage furniture, scissors, \\ pliers, safe, stapler} \\
         \midrule
         Test & \makecell[l]{blender, box, ladder, scale, dishwasher, \\ bin, chair} \\
    \bottomrule
    \end{tabular}
    \caption{Train-test split on object categories for fine-tuning LISA-13B.}
    \label{tab:lisa-train-test}
\end{table}

\begin{table}[t]
    \centering
    \begin{tabular}{ll}
    \toprule
        Number of Epoch & 10 \\
        Image Size & 1024 \\
        LoRa Rank & 8 \\
        LoRa Alpha & 16 \\
        LoRa Dropout & 0.05 \\
        Vision Encoder & CLIP-VIT-L \\
        Steps per Epoch & 500 \\
        Batch Size & 4 \\
        Optimizer & AdamW \\
        Learning Rate & 3e-4 \\
        Weight Decay & 0.0 \\
        Betas & [0.9, 095] \\
        Learning Rate Scheduler & WarmupDecayLR \\
        Warm up Steps & 100 \\
        Warm up Type & Linear \\
        Gradient Clipping & 1.0 \\
        $\lambda_{ce}$ & 1.0 \\
        $\lambda_{dice}$ & 0.5 \\
        $\lambda_{bce}$ & 2.0 \\
    \bottomrule
    \end{tabular}
    \caption{Hyper-parameters for fine-tuning LISA.}
    \label{tab:hyper-param}
\end{table}

\begin{table*}[t]
    \scriptsize
    \centering
    \begin{tabular}{lcccccccccccc}
    \toprule
    \multirow{2}{*}{Model} & \multicolumn{3}{c}{Lid} & \multicolumn{3}{c}{Blender} & \multicolumn{3}{c}{Washing} & \multicolumn{3}{c}{Box} \\
    \cline{2-13}
    & gIoU↑ & cIoU↑ & Sim↑ & gIoU↑ & cIoU↑ & Sim↑ & gIoU↑ & cIoU↑ & Sim↑ & gIoU↑ & cIoU↑ & Sim↑ \\
    \hline
    LISA-13B & 1.15 & 1.70 & 0.00 & 1.40 & 3.36 & 0.00 & 0.51 & 2.18 & 0.00 & 1.65 & 2.52 & 0.00 \\
    LISA-13B-Finetuned & 12.44 & 13.10 & 0.24 & 8.43 & 7.90 & 0.25 & 15.53 & 14.88 & 0.38 & 12.71 & 11.46 & 0.27 \\
    \hline
    \multirow{2}{*}{Model} & \multicolumn{3}{c}{Scale} & \multicolumn{3}{c}{Dishwasher} & \multicolumn{3}{c}{Bin} & \multicolumn{3}{c}{Chair} \\
    \cline{2-13}
    & gIoU↑ & cIoU↑ & Sim↑ & gIoU↑ & cIoU↑ & Sim↑ & gIoU↑ & cIoU↑ & Sim↑ & gIoU↑ & cIoU↑ & Sim↑ \\
    \hline
    LISA-13B & 0.61 & 1.41 & 0.00 & 0.00 & 0.00 & 0.00 & 5.92 & 10.97 & 0.00 & 0.05 & 0.06 & 0.00 \\
    LISA-13B-Finetuned & 14.32 & 13.51 & 0.25 & 9.56 & 8.86 & 0.15 & 3.55 & 6.09 & 0.10 & 6.05 & 4.33 & 0.55 \\
    \bottomrule
    \end{tabular}
    \caption{Comparison between LISA-13B and LISA-13B-Finetuned on the test set of EIVA15k.}
    \label{tab:lisa-compair}
\end{table*}

Our fine-tuning work was completely carried out on the basis of the official open sourced checkpoint and code repository of LISA-13B~\footnote{https://github.com/dvlab-research/LISA}.
We fine-tuned LISA-13B with a split of our EIVA15k by object categories.
The split is shown in table~\ref{tab:lisa-train-test}.
Contact area labels $Y_{contact} \in [0, 255]^{H\times W}$, which are heatmaps obtained by gaussian blur, is binarized using a threshold of 200.
To maintain the generalized ability of LISA, we also mixed the reasoning segmentation dataset used in LISA, and sample our affordance training data and reasoning segmentation data with a ratio of 1:1.

During fine-tuning, we optimize the same loss as the original loss proposed in LISA, which is
the combination of text generation loss $\mathcal{L}_{txt}$ and segmentation loss $\mathcal{L}_{mask}$:
\begin{equation}
    \mathcal{L} = \lambda_{txt} \mathcal{L}_{txt} + \lambda_{mask} \mathcal{L}_{mask}
\end{equation}
where $\lambda_{txt}$ and $\lambda_{mask}$ are balancing coefficients of these losses.
$\mathcal{L}_{txt}$ is the auto-regressive cross-entropy loss for text generation, and
$\mathcal{L}_{mask}$ is the combination of per-pixel binary cross-entropy (BCE) loss an DICE loss, with corresponding loss weights $\lambda_{bce}$ and $\lambda_{dice}$.
Hyper-parameters we used are summarized in Table~\ref{tab:hyper-param}.

Quantitative results of LISA-13B and LISA-13B-Finetuned on test set are demonstrated in Table~\ref{tab:lisa-compair}.
We observed a dramatic improvement on affordance prediction on the test set (i.e., unseen objects) after fine-tuned LISA-13B on our EIVA15k.
The results indicates that although LISA has a strong capability in solving general reasoning segmentation tasks, it lacks of the ability of inferring affordance.
This may due to the limitation of the foundation model (LLAMA2) used by LISA, beacause it does not well equipped with the world knowledge of manipulations.
However, LISA unlocked its ability of affordance prediction after fine-tuning on our EIVA15k, as indicated by the large improvement on both contact area prediction and motion direction prediction.
Such result backs up the idea of not only use EIVA15k as a benchmark, but also a high quality dataset for training.

\section{Performance Metrics}

\begin{enumerate}
\item \textbf{Similarity Metric (SIM):} SIM~\cite{DBLP:journals/ijcv/SwainB91} quantifies the similarity between the predicted affordance distribution and the ground truth. It is computed as the sum of the minimum values at each pixel location between the predicted contact region map and the ground-truth map.
\item \textbf{AUC-Judd (AUC-J):} AUC-J~\cite{DBLP:conf/iccv/JuddEDT09} is a variant of the AUC metric. This metric evaluates the proportion of the ground truth  captured by the predicted affordance map across different thresholds~\cite{DBLP:journals/pami/BylinskiiJOTD19}.
\item \textbf{Normalized Scanpath Saliency (NSS):} NSS~\cite{PETERS20052397} measures the correlation between the predicted affordance map and the ground truth. It is computed by normalizing the predicted contact region map to have zero mean and unit standard deviation, followed by averaging over ground-truth locations.
\item \textbf{Cosine Similarity (CS):} CS quantifies the directional similarity between the predicted and ground-truth vectors. It is calculated as the cosine of the angle between these two vectors, normalized to a range of [-1, 1], where 1 indicates perfect alignment, 0 indicates orthogonality, and -1 represents complete opposition. An acceptable result is that this metric is at least greater than 0, indicating that these two vectors are aligned in the same direction rather than being opposite.
\end{enumerate}

\end{document}